\documentclass{article}

\usepackage{arxiv}

\usepackage[utf8]{inputenc} 
\usepackage[T1]{fontenc}    
\usepackage{hyperref}       
\usepackage{url}            
\usepackage{booktabs}       
\usepackage{amsfonts}       
\usepackage{nicefrac}       
\usepackage{microtype}      
\usepackage{lipsum}		
\usepackage{graphicx}
\usepackage{natbib}
\usepackage{doi}
\usepackage{wrapfig}

\title{Breaking the attention bottleneck}

\author{
  Kalle Johannes Hilsenbek \\
	\texttt{s2kahils@uni-trier.de} \\
}

\renewcommand{\headeright}{GNU AGPL v3 license}
\renewcommand{\undertitle}{Global context as static attention replacement in decoders}
\renewcommand{\shorttitle}{Breaking the attention bottleneck}

\hypersetup{
pdftitle={Breaking the attention bottleneck},
pdfsubject={nlp},
pdfauthor={Kalle Johannes Hilsenbek},
pdfkeywords={linear attention, transformer, decoder, generative language modeling, GPT},
}

\begin{document}
\maketitle

\begin{abstract}
Attention-based transformers have become the standard architecture in many deep learning fields, primarily due to their ability to model long-range dependencies and handle variable-length input sequences.
However, the attention mechanism with its quadratic complexity is a significant bottleneck in the transformer architecture. This algorithm is only uni-directional in the decoder and converges to a static pattern in over-parametrized decoder-only models.
I address this issue by developing a generative function as attention or activation replacement.
It still has the auto-regressive character by comparing each token with the previous one.
In my test setting with nanoGPT this yields a smaller loss while having a smaller model.
The loss further drops by incorporating an average context vector.
This concept of attention replacement is distributed under the GNU AGPL v3 license at \url{https://gitlab.com/Bachstelze/causal_generation}.
\end{abstract}


\section{Introduction}

The advent of attention-based transformer models has catalyzed transformative advancements in the fields of natural language processing, video, and audio processing. Since their introduction, these models have consistently pushed the boundaries of what is achievable in machine learning and artificial intelligence. The work on the Transformer architecture by Vaswani et al. \cite{vaswani2023attention} laid the foundation for a new era in which attention mechanisms enable models to process and understand complex sequential data with an improved accuracy.

\begin{wrapfigure}{r}{0.32\textwidth} 
    \centering
    \includegraphics[width=0.32\textwidth]{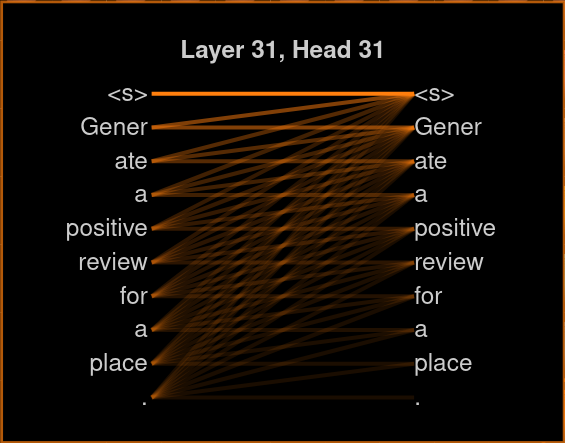}
    \caption{This attention pattern is repeated in all heads of instruction-tuned GPTs. Each token has a uniform distribution over all previous tokens as a consistent auto-regressive activation.}
    \label{static_pattern}
\end{wrapfigure}
One-directional attention models, such as GPT\cite{brown2020language}, are commonly employed for instruction-following tasks, highlighting the significance of model scaling in the development of decoder-only architectures. However, this increase of the decoder size leads to computational demands, a "lost-in-the-middle"-effect of accessible input informations \cite{liu2023lost} and an informationless redundant attention pattern (see figure \ref{static_pattern} on the right) \cite{explainable_attention}.
Hence, instruction-tuned GPTs only store information in the fully connected linear feed-forward layers. They model information mostly linear \cite{hernandez2024linearity}\cite{razzhigaev2024transformer} and represent abstract concepts linearly \cite{park2023linear}. Therefore, instruction GPTs can be seen as advanced multi-layer perceptrons and generative key-value memories \cite{geva2021transformer}.
The paper "Transformers are Multi-State RNNs" proposes that transformers can be viewed as a type of recurrent neural network (RNN) with multiple states \cite{oren2024transformers}.
Traditionally, transformers and RNNs are seen as distinct architectures. Transformers rely on self-attention mechanisms to handle dependencies between input tokens, while RNNs auto-regressive process sequences step-by-step. However, transformers can be interpreted as multi-state RNNs, where each layer in a transformer manages a set of hidden states, much like an RNN but with multiple-state representations.
\newpage
While the quadratic complexity of attention is clobbered by large tech companies with a destructive amount of resources, it poses significant challenges for the broader research community.
Therefore, the resource-intensive nature of these models has prompted the exploration of more efficient transformer variants.
For example, "Your Transformer is Secretly Linear" reveals a novel linear characteristic in transformer decoders, suggesting that their operation may be more linear than previously assumed.
This study introduces a cosine-similarity-based regularization to reduce layer linearity, improving performance metrics on benchmarks like Tiny Stories and SuperGLUE \cite{razzhigaev2024transformer}.
In contrast, "Not All Language Model Features Are Linear" argues that certain features of language models resist linear simplification, highlighting the need for a nuanced approach to optimization \cite{engels2024language}.

In this paper, I try to explore the limitations and challenges associated with current transformer models, particularly focusing on the "attention bottleneck". This bottleneck arises from the quadratic complexity of the attention mechanism, which can hinder the scalability and efficiency of transformer models when processing long sequences. I will examine simple techniques aimed at breaking this bottleneck, paving the way for more efficient and scalable transformer-based solutions. By addressing these challenges, I aim to contribute to the ongoing evolution of transformer models, ensuring their accessibility and efficacy within the open-source community.

\section{Related work}

In the field of language modeling, linear attention mechanisms have been developed to address quadratic complexity and keep the dynamic parameterized behavior\cite{wu2021fastformer}\cite{katharopoulos2020transformers}\cite{schlag2021linear}\cite{zhai2021attention}\cite{nguyen2022momentum}\cite{arora2023zoology}\cite{lee2024sea} \cite{peng2023rwkv}\cite{peng2024eagle}. These methods aim to make attention computation more efficient in terms of both memory and computational resources, allowing transformers to handle longer sequences effectively \cite{buckman2024}.
E.g. by representing keys and queries with kernel feature maps, it enables the use of associative properties to simplify the computation \cite{shen2024efficient} \cite{katharopoulos2020transformers}.
The Performer uses kernel methods to approximate the softmax function \cite{choromanski2022rethinking}.
New approaches have further optimized the computation by employing kernel tricks for different memory components within GPUs to enhance execution speed and efficiency \cite{qin2024lightning}\cite{sun2024linear}.
Furthermore, a linear transformation by an unparameterized Fourier Transform can encode language with a reduction in accuracy \cite{leethorp2022fnet}.
There are many more attempts to optimize attention besides its linearization \cite{tay2022efficient}\cite{hosseini2024need}\cite{munkhdalai2024leave}.
E.g. FlashAttention is introduced as a new standard method to compute attention exactly and efficiently. The method is designed to address the inefficiencies in memory and operations from the input to the output that are common in traditional attention mechanisms \cite{dao2022flashattention}.

There are also parameter-free attention approaches with convolutional neural nets (CNN) in vision processing
\cite{körber2022parameterfree} \cite{shi2023parameterfree} \cite{pmlr-v139-yang21o} \cite{guo2022calip} \cite{wan2023swift}.
While they don't explicitly delve into language modeling, the concepts of parameter-free attention mechanisms have relevant implications for this domain. Similar to how they aim to simplify attention in CNNs, parameter-free attention mechanisms in language models can reduce the computational load and memory requirements. This makes training and inference more efficient, particularly beneficial for large-scale and resource-intensive language models.

\section{Method}

In this section, I describe the cross-entropy loss divergence as a measurement between the standard attention and my concept of recurrence and global context as static attention replacement in nanoGPT. A small setting is used with a block size of 64, a batch size of 12, 4 layers, an embedding size of 128, no dropout, and 4 heads for attention. The only difference to the hyperparameters in the ReadMe is the longer iteration and learning rate decay iteration number of 5000 instead of 2000 to see that the results also hold for a longer training\cite{nanoGPT}.
The small nanoGPT configuration has a number of 0.8 million parameters. Those parameters accumulate from 802,944 decayed parameter tensors and 1,152 non-decayed parameter tensors. This amount is reduced to 0.6 million without the attention parameters of the key, value, and query matrices. The quarter smaller size comes from the total 606,336 parameters of decayed parameter tensors in my  static attention approach.
A comparison of the different concepts is made by distinguishing the loss curve of the training on the tiny Shakespeare dataset \cite{shakespear}.

This Shakespeare dataset is so small that the middle model size of nanoGPT overfits with 10.65 million number of parameters. The middle-sized setting has 6 layers, 6 attention heads, an embedding size of 384, batch size of 64 and also a learning-rate of 1e-3.
My concept is also applied in this middle-sized setting to evaluate its performance in an over-parameterized setting.
\newpage
\section{Results}

\begin{figure}[h]
      \centering
      \includegraphics[width=1\linewidth]{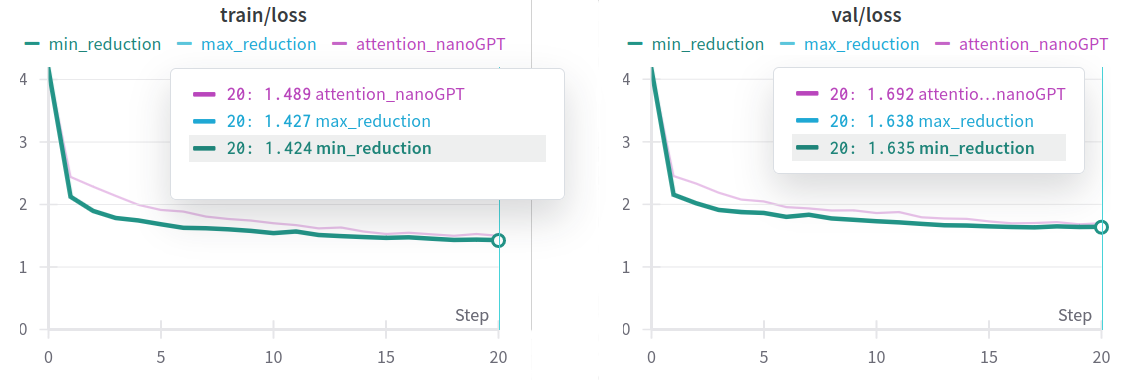}
      \caption{The training and validation loss curve of the standard attention and recurrent maximization as well as minimization.}
      \label{nanoGPT_loss}
\end{figure}

The auto-regressive attention can be replaced by static functions without parameters by mimicking its equal distribution over the previous tokens (see figure \ref{static_pattern} in the introduction). This can achieved by returning the maximum value of all dimensions of the current token and its precursor. The first token remains unchanged by this vector manipulation. Such a simple concept already achieves a smaller loss value with a smaller model size at once, while reducing the computational cost. This finding is significant at the start of the training and aligns with the further process (see figure \ref{nanoGPT_loss}). Causal attention has a validation loss of 1.692, whereas my approach has a difference of 0.054 with a loss of 1.638 in the end. Minimizing with the previous token yields a nearly identical loss of 1.635.
Training fluctuation could be a plausible explanation for the tiny loss difference between maximization and minimization with the previous token.

\begin{figure}[h]
      \centering
      \includegraphics[width=1\linewidth]{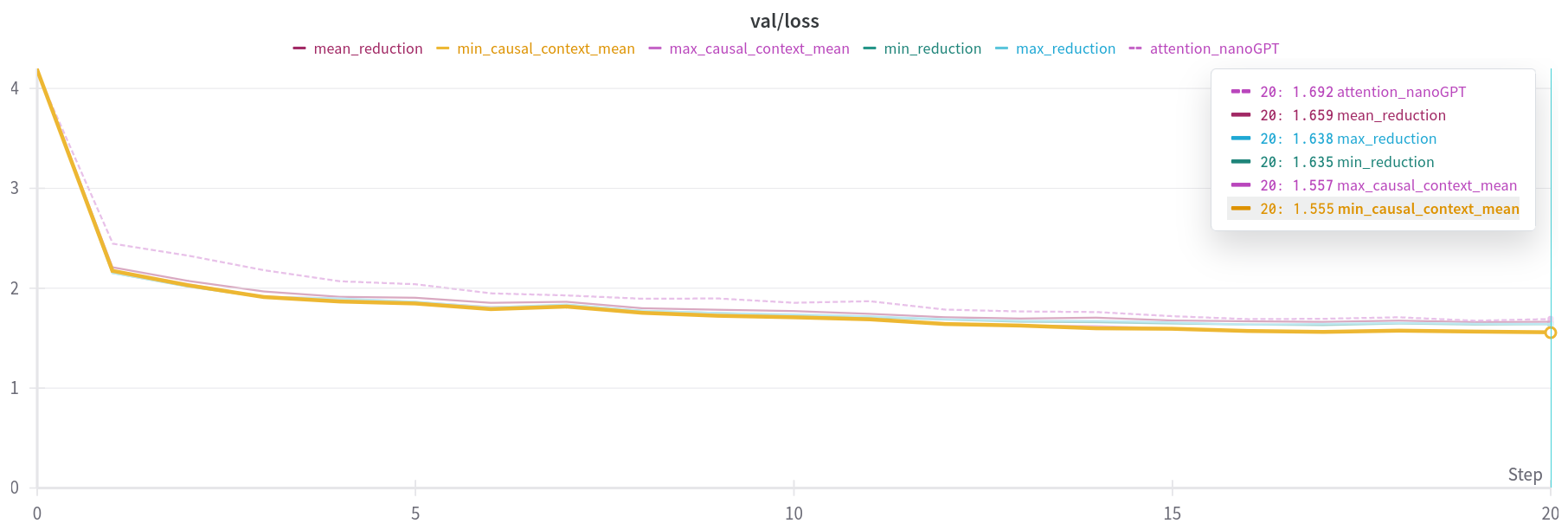}
      \caption{The validation loss curve of the standard attention and recurrent methods with an average context.}
      \label{nanoGPT_context_validation_loss}
\end{figure}

The integration of an average context vector as reference yields an even further improvement of the auto-regressive maximization loss down to 1.557 respectively to 1.555 for the minimization. This drop of around 0.135 from the standard attention loss is attained by the additional averaging of all input vectors and an additional comparison of all tokens to this reference vector. Calculating the mean between vectors seems to signal a complementary signal to the maximization or minimization. Only calculating the mean with the previous token performs between the standard attention and the discrete auto-regressive optimization (see figure \ref{nanoGPT_context_validation_loss}).

\begin{figure}[h]
      \centering
      \includegraphics[width=1\linewidth]{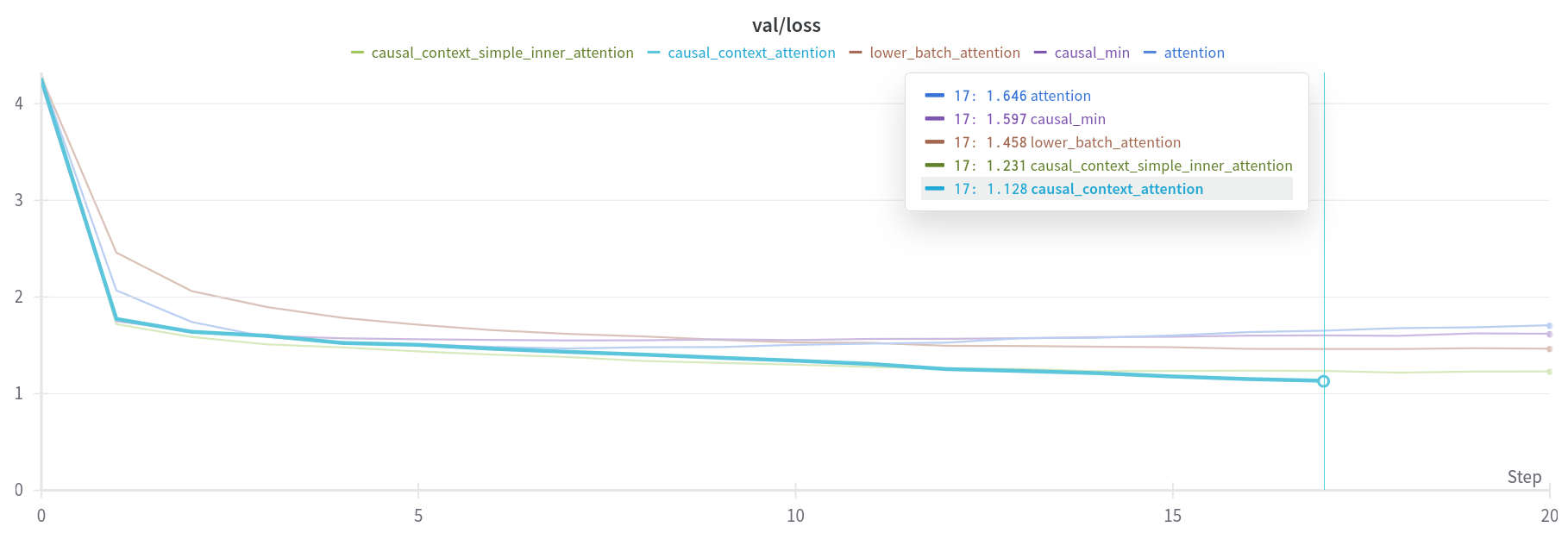}
      \caption{The validation loss curve of the standard attention and recurrent methods with an average context in the middle Shakespeare setting of nanoGPT.}
      \label{validation_gpu_train}
\end{figure}

In the over-parameterized setting the validation loss curve of the standard attention increases after a fast drop of the validation the loss to a local optimum. This can be mitigated by reducing the batch size from 64 to 16, which yields a slightly better validation loss for attention (see figure \ref{validation_gpu_train}) that is continuously decreasing.
Replacing the attention module with my recursive function shows a similar over-fitting behaviour (see the curve of "causal\_min" in  figure \ref{validation_gpu_train}).
This over-fitting can be avoided by leveraging my generative function as activation in the fully-connected feedforward network. This complementary approach stabilized and improves the training even with the bigger batch size of 64 (see the "causal\_context\_attention"\-curve in  figure \ref{validation_gpu_train}). Though the activation replacement increases the computational cost from 500ms to around 575ms per iteration and the 2.5GB GPU-RAM to 3.6GB. This increase is caused by the 4 times up-scaling of the inner dimension of the perceptron. The replaced GELU function is activated before the down-scaling the normal embedding size.
Leaving the inner dimension unchanged reduces the cost of the generative activation to 2.2GB GPU-RAM and an iteration time of 325ms while having similar results to the upscaled version (see the "causal\_context\_simple\_inner\_attention"-curve in  figure \ref{validation_gpu_train}). My generative activation can also be placed before the first linear computation to keep the larger inner dimensions with the GELU-activation.

\section{Disussion}
In "Multilayer feedforward networks are universal approximators" it is demonstrated that perceptrons can approximate any continuous function\cite{HORNIK1989359}\cite{Approximation}. This result is often referred to as the Universal Approximation Theorem and it could also describe the generative pretrained transformers. The feedforward neural network seems to be the crucial component within the GPT architecture by considering the static help function of attention in decoders.\footnote{There is a discussion in BertViz about this redundant pattern: \url{https://github.com/jessevig/bertviz/issues/128}} Therefore it should be possible to apply all of the Universal Approximation Theorem to decoder-only models.
Attention can be seen as an activation function for the perceptron \cite{neo2024interpreting}.
Non-linear activation functions (like static attention) enable the network to approximate complex functions.
The generative power comes with such non-linearity and model size.

An open question is the combination of the different algorithm variants at many possible positions and furthermore the integration of my concept into other architectures like encoder-decoder models or plain MLPs \cite{10.5555/944919.944966}. Moreover, the context vector could be gated with external resources, e.g. for nearest neighbor speculative decoding\cite{li2024nearest}. This could enhance the model's ability to incorporate external knowledge dynamically, improving its overall contextual understanding and decision-making capabilities while keeping a low computational complexity.

\section{Conclusion}

In this paper, I presented an efficient attention replacement for the decoder module that addresses some of the key challenges faced by traditional attention mechanisms. My proposed method can significantly reduce the computational complexity and resource requirements while maintaining or even improving the language modeling. The amount of operations is linear with the sequence length and increases by a logarithmic factor with the self context vector. The complexity can be brought down to O(1) and O(log(n)) for the averaging over all tokens in parallel. This makes it particularly suitable for deployment in environments with limited computational resources and hopefully helps to reduce the destructive impact on our nature.

By leveraging a generative help function, my approach not only simplifies the model architecture but also enhances interpretability and transparency. This aligns with the growing demand for explainable AI systems that can provide insights into their decision-making processes. The empirical results demonstrate so far that my method achieves competitive modeling, which still has to be proven in standard settings on downstream tasks.

I am pleased to distribute this work under the GNU affero general public lisence version 3 ("GNU AGPL v3 license") at \url{https://gitlab.com/Bachstelze/causal_generation} to ensure that the research community can freely access, combine and build upon my findings. This open-access approach fosters collaboration and innovation, encouraging a development, which is aligned with humanistic values.

\bibliographystyle{unsrtnat}
\bibliography{references}  






\end{document}